\documentclass[sigconf]{acmart}
\AtBeginDocument{%
  }

\copyrightyear{2025}
\acmYear{2025}
\setcopyright{rightsretained}
\acmConference[ICAIL 2025]{20th International Conference on Artificial Intelligence and Law}{June 16--20, 2025}{Chicago, IL, USA}
\acmBooktitle{20th International Conference on Artificial Intelligence and Law (ICAIL 2025), June 16--20, 2025, Chicago, IL, USA}
\acmPrice{}
\acmDOI{10.1145/3769126.3769200}
\acmISBN{979-8-4007-1939-4/25/06}

\usepackage{scalerel}
\usepackage{tikz}
\usetikzlibrary{svg.path}

\definecolor{orcidlogocol}{HTML}{A6CE39}
\tikzset{
  orcidlogo/.pic={
    \fill[orcidlogocol] svg{M256,128c0,70.7-57.3,128-128,128C57.3,256,0,198.7,0,128C0,57.3,57.3,0,128,0C198.7,0,256,57.3,256,128z};
    \fill[white] svg{M86.3,186.2H70.9V79.1h15.4v48.4V186.2z}
                 svg{M108.9,79.1h41.6c39.6,0,57,28.3,57,53.6c0,27.5-21.5,53.6-56.8,53.6h-41.8V79.1z M124.3,172.4h24.5c34.9,0,42.9-26.5,42.9-39.7c0-21.5-13.7-39.7-43.7-39.7h-23.7V172.4z}
                 svg{M88.7,56.8c0,5.5-4.5,10.1-10.1,10.1c-5.6,0-10.1-4.6-10.1-10.1c0-5.6,4.5-10.1,10.1-10.1C84.2,46.7,88.7,51.3,88.7,56.8z};
  }
}

\newcommand\orcidicon[1]{\href{https://orcid.org/#1}{\mbox{\scalerel*{
\begin{tikzpicture}[yscale=-1,transform shape]
\pic{orcidlogo};
\end{tikzpicture}
}{|}}}}

\usepackage{hyperref} %<--- Load after everything else

\begin{document}

\title{Data Augmented Pipeline for Legal Information\\Extraction and Reasoning}

\author{Nguyen Minh Phuong}
\orcid{0000-0002-3752-8699}
\affiliation{%
  \institution{Center for Juris-Informatics, ROIS-DS, \\Tokyo, Japan}
  \city{}
  \country{}
}
\affiliation{%
  \institution{Japan Advanced Institute of Science and Technology, Ishikawa, Japan}
  \city{}
  \country{}
}
\email{phuongnm@jaist.ac.jp}

\author{Ha-Thanh Nguyen}
\orcid{0000-0003-2794-7010}
\affiliation{%
  \institution{Center for Juris-Informatics, ROIS-DS, \\Tokyo, Japan}
  \city{}
  \country{}
}
\affiliation{%
  \institution{Research and Development Center for LLMs, NII, \\Tokyo, Japan}
  \city{}
  \country{}
}
\email{nguyenhathanh@nii.ac.jp}

\author{May Myo Zin}
\orcid{0000-0003-1315-7704}
\affiliation{%
  \institution{Center for Juris-Informatics, ROIS-DS, \\Tokyo, Japan}
  \city{}
  \country{}
}
\email{maymyozin@nii.ac.jp}

\author{Ken Satoh}
\orcid{0000-0002-9309-4602}
\affiliation{%
  \institution{Center for Juris-Informatics, ROIS-DS, \\Tokyo, Japan}
  \city{}
  \country{}
}
\email{ksatoh@nii.ac.jp}

\renewcommand{\shortauthors}{Phuong et al.}

\begin{abstract}
In this paper, we propose a pipeline leveraging Large Language Models (LLMs) for data augmentation in Information Extraction tasks within the legal domain. The proposed method is both simple and effective, significantly reducing the manual effort required for data annotation while enhancing the robustness of Information Extraction systems. Furthermore, the method is generalizable, making it applicable to various Natural Language Processing (NLP) tasks beyond the legal domain.
\end{abstract}





\maketitle
\section{Introduction} The PROLEG knowledge representation language \citep{Satoh2023} is designed to help lawyers engage with legal reasoning systems. While it does not address all challenges, such as limited expressiveness for certain legal concepts and the ambiguity of legal texts, it provides a minimal yet sufficient language for reasoning, enabling lawyers to understand system behavior. Initially, given a legal contract or agreement, the Deep PROLEG system \citep{Satoh2023,deep_proleg_demo} is equipped with a set of rules in the form of Horn clauses accompanied by a set of exception expressions \citep{Satoh2023}.
% todo: related work related to symbolic reasoner recently 
Subsequently, a deep neural semantic parser is trained to transform legal cases into a set of fact expressions, which are then used to determine entailment with the installed contract within the Deep PROLEG system. However, extending the system to new domains requires substantial annotated data, which is both time-consuming and labor-intensive. Moreover, legal experts often lack expertise in NLP and face difficulties in mastering state-of-the-art machine learning techniques. To address these challenges, this work proposes a semi-supervised method that integrates new knowledge into the model through data-augmented samples, while also leveraging LLMs to address domain-specific tasks.
% An overview of our pipeline within the overall Deep PROLEG system is presented in Figure~\ref{fig_overview}.
    
\begin{figure*}
    \centering
    \includegraphics[width=0.9\textwidth, keepaspectratio, 
            trim={0.cm 0.38cm 0.cm  0.37cm}, page=1, clip=true]{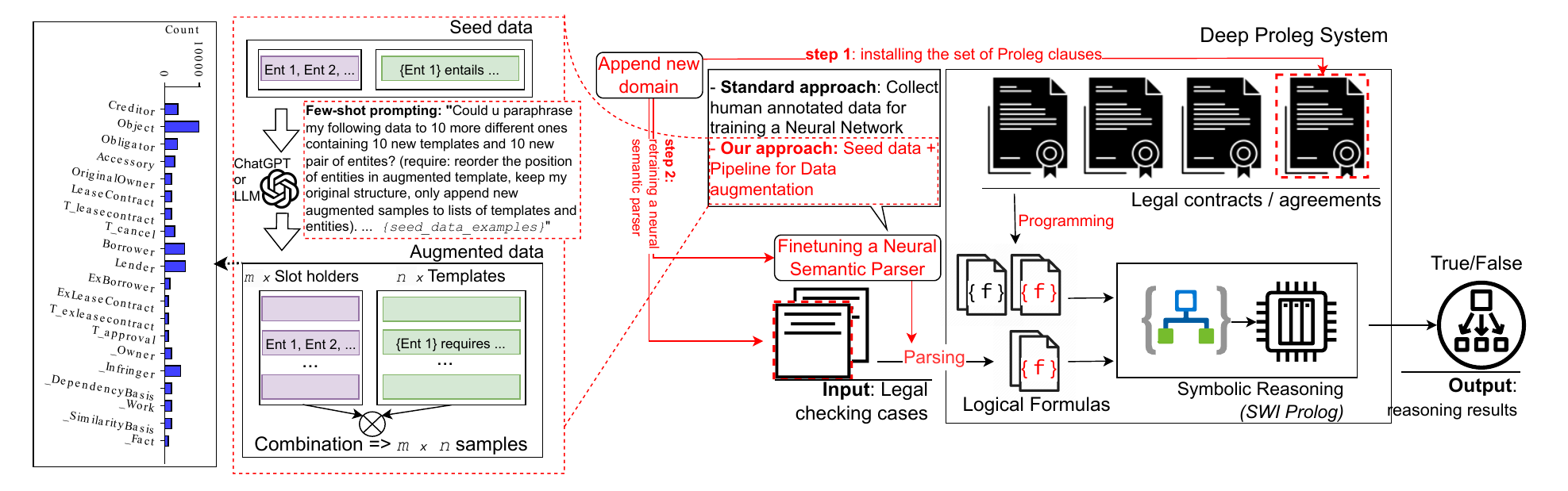} 
    \caption{The general architecture of the deep PROLEG system is depicted, along with the distribution of legal slot holders in the augmented data. The components related to the process of adding a new domain are highlighted in \textcolor{red}{red}.}
    \label{fig_overview}
\end{figure*}

% \vspace{-5pt}
\section{System Description}
In this work, we aim to design a mechanism that promptly extends the Deep PROLEG system \citep{Satoh2023,deep_proleg_demo} to various domains with less human effort. We first introduce the Deep PROLEG system, and then our pipeline is integrated into this system for a new domain adaptation. 
\subsection{Overall Deep PROLEG System} 
In general, the Deep PROLEG system comprises three major modules: (1) the Natural Language Perceiver - a neural semantic parser receives a legal case and parses it into the facts in legal knowledge representation language; (2) the PROLEG Reasoner - the logical rules of all legal contracts or agreements are installed in a Symbolic Reasoner using the PROLEG language. The facts that were obtained from the legal cases outputted in the previous step are transferred to this symbolic reasoner to verify the truth value of the goal expression; and (3) the Inference Explainer - the module tracks the logical inference in the symbolic reasoner and visualizes the inferencing flow, which supports inspecting the reasoning process.
% \begin{itemize}
%     \item Natural Language Perceiver: 
%     \item PROLEG Reasoner: In this component, 
%     \item Inference Explainer: This . 
% \end{itemize}
Finally, the Deep PROLEG system assists lawyers or courts in evaluating legal cases alongside contract documents to swiftly determine the entailment of these cases with the established contracts. Additionally, the system outputs detailed results of the inference flow and corresponding truth values for each reasoning step. 

\noindent\textbf{Challenges of Scaling Up.} The process of adding a new contract to the Deep PROLEG system involves two main steps (Figure~\ref{fig_overview}): (1) manually installing the set of PROLEG clauses (grounding) that are implied in the contract and (2) retraining a neural semantic parser model to convert a legal case query into a set of facts that support logical inference. The second step is labor-intensive and time-consuming for experts. Moreover, prior data augmentation methods using heuristics \citep{jia-liang-2016-data} or syntactic cues \citep{sutiono-hahn-powell-2022-syntax} lack generalization to complex legal cases. This work proposes a simple yet effective approach using few-shot prompting and LLMs \citep{NEURIPS2020_1457c0d6}, generating augmented data to train a new neural semantic parser informed by new contracts and agreements.
\subsection{Pipeline for Data Augmentation}
\label{sec_pipeline_domain_adaptaion}
In this section, we provide a comprehensive description of our approach to the data augmentation method. Drawing inspiration from the distinctions between \textit{function} and \textit{content} words, we posit that legal cases can be deconstructed into two types of information: \textit{templates} and \textit{slot holders}. In essence, each combination of a set of slot holders and a template produces a single legal case sample. A slot holder may represent an entity name or a text span with specific significance (Table~\ref{tab_example}). 
\begin{table}[]
    \small
    \centering
    \caption{  Examples of Augmented Templates and Slot Holders \vspace{-16pt}
    \label{tab_example} }
    \resizebox{0.49\textwidth}{!}{%
    \begin{tabular}{lp{0.65\textwidth}}
    \toprule
     Slot holders 1\vspace{2pt}&  
   \{"Object":"the house","Accessory":"garage A","OriginalOwner":"sarah","Creditor":"john","Obligator":"alex"\} \\\vspace{1pt}
     Slot holders 2&   \{"Object":"the apartment",
   "Accessory": "balcony C", ... \}  \\\midrule
     Template 1 \vspace{2pt}& After \{OriginalOwner\} inherited \{Object\} from \{Creditor\}, \{Creditor\} came across \{Obligator\} at \{Object\}, who had erected \{Accessory\}. \{Creditor\} requested \{Obligator\} to leave \{Object\} and dismantle \{Accessory\}. In response, \{Obligator\} asserted that they rented \{Object\} from \{OriginalOwner\}, thus claiming rights over \{Accessory\}. Will \{Creditor\} be able to reclaim \{Object\}?  \\ 
     Template 2& During a visit to \{Object\}, \{Creditor\} discovered \{Obligator\} residing there and having constructed \{Accessory\}, which was inherited ...  \\  \midrule
     PROLEG facts& \texttt{original\_ownership(\{OriginalOwner\},\{Object\}).}\\
        &\texttt{transfer(\{OriginalOwner\},\{Creditor\},\{Object\}).} \texttt{occupancy(\{Obligator\},\{Object\}).}\\
        &\texttt{existence\_of\_accessory(\{Accessory\},\{Object\}).} ... \\
     \bottomrule\vspace{-20pt}
    \end{tabular}
    }
\end{table} 
Utilizing the few-shot prompting technique \citep{NEURIPS2020_1457c0d6}, we manually create one or two templates and entities as seed data, which then guide LLMs to generate additional templates and sets of entities, as depicted in Figure~\ref{fig_overview}. In addition, the corresponding PROLEG facts of these augmented samples can also be generated based on the set of provided entities. Then, the augmented data will be aggregated to yield legal case samples, slot information as entities, and the respective PROLEG legal facts implied in augmented samples. This augmented data is used to fine-tune a Neural Semantic Parser model for new domain adaptation.
\paragraph{Neural Semantic Parsing} Building upon prior work \citep{deep_proleg_demo}, our system was implemented and evaluated using two approaches: an end-to-end machine translation model and a named entity recognition (NER)-based model. In the first approach, the semantic parser directly produces the final logical forms (facts). Conversely, in the NER-based approach, we utilize a simplified look-up table derived from the training data to infer predicates based on the set of recognized slot holders (or entities). 

\subsection{Results and Performance Evaluation}
Our proposed method significantly reduces the human effort required to implement new legal domains from scratch. The experiments employed ChatGPT for data augmentation, including GPT-3.5 Turbo, GPT-4o mini, and GPT-4o. Preliminary tests with open-source LLMs such as Qwen2.5-14B and Meta-Llama-3-8B produced high-quality data but did not outperform ChatGPT. To this end, the Deep PROLEG system is working effectively with the performance of more than 95\% accuracy over the augmented dataset containing 5000 legal cases within four kinds of contracts and 20 different legal slot holders (distribution is depicted in Figure~\ref{fig_overview}).  
% \begin{figure}
%     \centering
%     \includegraphics[width=0.5\textwidth,keepaspectratio, 
%             trim={0 0.5cm 0 0}, page=1, clip=true]{images/0a.pdf}\vspace{-5pt}
%     \caption{The distribution of legal slot holders.
%     \vspace{-4pt}
%     }
%     \label{fig_distribution}
% \end{figure}

% \begin{itemize}
% \item number of training/test samples 
% \item number of templates + samples 
% \item model performance between 2 approaches: NER (single and separated models) + machine translation
% % \item 
% \end{itemize}
\section{Conclusion}
In this work, we present a practical framework for data augmentation methods aimed at rapidly adapting the Deep PROLEG system to new legal domains. Our method is simple yet effective, significantly reducing the human effort required for creating annotated data. We believe that our augmentation method can be widely applied to other NLP domains, particularly within the information extraction or semantic parsing areas.

% \section{SIGCHI Extended Abstracts}

% The ``\verb|sigchi-a|'' template style (available only in \LaTeX\ and
% not in Word) produces a landscape-orientation formatted article, with
% a wide left margin. Three environments are available for use with the
% ``\verb|sigchi-a|'' template style, and produce formatted output in
% the margin:
% \begin{description}
% \item[\texttt{sidebar}:]  Place formatted text in the margin.
% \item[\texttt{marginfigure}:] Place a figure in the margin.
% \item[\texttt{margintable}:] Place a table in the margin.
% \end{description}

% \begin{acks}
% \end{acks}

\bibliographystyle{ACM-Reference-Format}
\bibliography{ref}

% \appendix



\end{document}